\documentclass[10pt,twocolumn,letterpaper]{article}

\usepackage{cvpr}
\usepackage{times}
\usepackage{epsfig}
\usepackage{graphicx}
\usepackage{amsmath}
\usepackage{amssymb}
\usepackage{verbatim}

\usepackage{url}            
\usepackage{booktabs}       
\usepackage{amsfonts}       
\usepackage{nicefrac}       
\usepackage{microtype}      
\usepackage{xcolor}
\usepackage{adjustbox}
\usepackage{multirow}
\usepackage{floatrow}
\usepackage{subcaption}
\newfloatcommand{figurebox}{figure}[\nocapbeside][\dimexpr(\textwidth-\columnsep)/2\relax]
\newfloatcommand{tablebox}{table}[\nocapbeside][\dimexpr(\textwidth-\columnsep)/2\relax]

\usepackage[pagebackref=true,breaklinks=true,letterpaper=true,colorlinks,bookmarks=false]{hyperref}
\cvprfinalcopy 

\newcommand{\vv}[1]{\mathbf{#1}}

\def\cvprPaperID{435} 

\ifcvprfinal\fi
\begin{document}

\title{Deep Layer Aggregation}

\author{Fisher Yu \quad Dequan Wang \quad Evan Shelhamer \quad Trevor Darrell \vspace{1.5mm} \\ 
UC Berkeley \vspace{1mm} }

\maketitle

\begin{abstract}
    Visual recognition requires rich representations that span levels from low to high, scales from small to large, and resolutions from fine to coarse.
    Even with the depth of features in a convolutional network, a layer in isolation is not enough: compounding and aggregating these representations improves inference of what and where.
    Architectural efforts are exploring many dimensions for network backbones, designing deeper or wider architectures, but how to best aggregate layers and blocks across a network deserves further attention.
    Although skip connections have been incorporated to combine layers, these connections have been ``shallow''  themselves, and only fuse by simple, one-step operations.
    We augment standard architectures with deeper aggregation to better fuse information across layers.
    Our deep layer aggregation structures iteratively and hierarchically merge the feature hierarchy to make networks with better accuracy and fewer parameters.
    Experiments across architectures and tasks show that deep layer aggregation improves recognition and resolution compared to existing branching and merging schemes.
\end{abstract}


\section{Introduction}  

Representation learning and transfer learning now permeate computer vision as engines of recognition.
The simple fundamentals of compositionality and differentiability give rise to an astonishing variety of deep architectures \cite{krizhevsky2012imagenet,szegedy2015going,simonyan2015very,he2016deep,zagoruyko2016wide}.
The rise of convolutional networks as the backbone of many visual tasks, ready for different purposes with the right task extensions and data \cite{girshick2015region-based, shelhamer2016fully, xie2015holistically}, has made architecture search a central driver in sustaining progress.
The ever-increasing size and scope of networks now directs effort into devising design patterns of modules and connectivity patterns that can be assembled systematically.
This has yielded networks that are deeper and wider, but what about more closely connected?

More nonlinearity, greater capacity, and larger receptive fields generally improve accuracy but can be problematic for optimization and computation.
To overcome these barriers, different blocks or modules have been incorporated to balance and temper these quantities, such as bottlenecks for dimensionality reduction \cite{lin2014network, szegedy2015going, he2016identity} or residual, gated, and concatenative connections for feature and gradient propagation \cite{he2016identity, srivastava2015highway, huang2016densely}.
Networks designed according to these schemes have $100+$ and even $1000+$ layers.

\begin{figure}[tb]
\centering
\includegraphics[width=\linewidth]{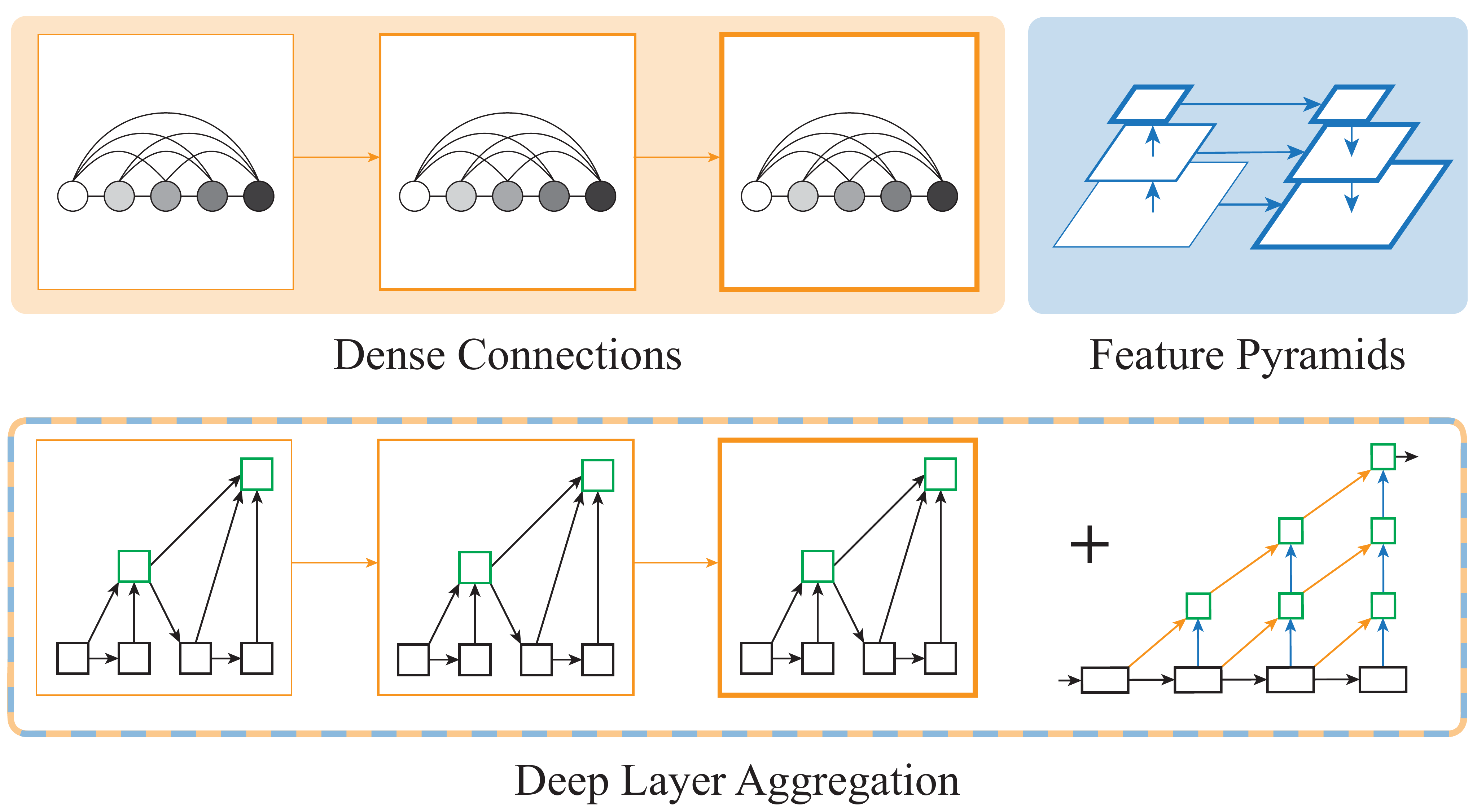}
\caption{
Deep layer aggregation unifies semantic and spatial fusion to better capture what and where.
Our aggregation architectures encompass and extend densely connected networks and feature pyramid networks with hierarchical and iterative skip connections that deepen the representation and refine resolution. 
}
\label{fig:whatwhere}
\end{figure}

Nevertheless, further exploration is needed on how to connect these layers and modules.
Layered networks from LeNet \cite{lecun1998gradient} through AlexNet \cite{krizhevsky2012imagenet} to ResNet \cite{he2016identity} stack layers and modules in sequence.
Layerwise accuracy comparisons~\cite{donahue2014decaf,zeiler2014visualizing,shelhamer2016fully}, transferability analysis~\cite{yosinski2014transferable}, and representation visualization~\cite{zeiler2014visualizing,yu2017dilated} show that deeper layers extract more semantic and more global features, but these signs do not prove that the last layer is the ultimate representation for any task.
In fact, skip connections have proven effective for classification and regression \cite{huang2016densely, bishop2006pattern} and more structured tasks \cite{hariharan2015hypercolumns, shelhamer2016fully, lin2016feature}.
Aggregation, like depth and width, is a critical dimension of architecture.


In this work, we investigate how to aggregate layers to better fuse semantic and spatial information for recognition and localization.
Extending the ``shallow'' skip connections of current approaches, our aggregation architectures incorporate more depth and sharing.
We introduce two structures for deep layer aggregation (DLA): iterative deep aggregation (IDA) and hierarchical deep aggregation (HDA).
These structures are expressed through an architectural framework, independent of the choice of backbone, for compatibility with current and future networks.
IDA focuses on fusing resolutions and scales while HDA focuses on merging features from all modules and channels.
IDA follows the base hierarchy to refine resolution and aggregate scale stage-by-stage.
HDA assembles its own hierarchy of tree-structured connections that cross and merge stages to aggregate different levels of representation.
Our schemes can be combined to compound improvements.

Our experiments evaluate deep layer aggregation across standard architectures and tasks to extend ResNet ~\cite{he2016deep} and ResNeXt~\cite{xie2016aggregated} for large-scale image classification, fine-grained recognition, semantic segmentation, and boundary detection.
Our results show improvements in performance, parameter count, and memory usage over baseline ResNet, ResNeXT, and DenseNet architectures.
DLA achieve state-of-the-art results among compact models for classification.
Without further architecting, the same networks obtain state-of-the-art results on several fine-grained recognition benchmarks.
Recast for structured output by standard techniques, DLA achieves best-in-class accuracy on semantic segmentation of Cityscapes \cite{cordts2016cityscapes} and state-of-the-art boundary detection on PASCAL Boundaries \cite{premachandran2017pascal}.
Deep layer aggregation is a general and effective extension to deep visual architectures.

\section{Related Work}

We review architectures for visual recognition, highlight key architectures for the aggregation of hierarchical features and pyramidal scales, and connect these to our focus on deep aggregation across depths, scales, and resolutions.

The accuracy of AlexNet \cite{krizhevsky2012imagenet} for image classification on ILSVRC \cite{russakovsky2015imagenet} signalled the importance of architecture for visual recognition.
Deep learning diffused across vision by establishing that networks could serve as backbones, which broadcast improvements not once but with every better architecture, through transfer learning \cite{donahue2014decaf, zeiler2014visualizing} and meta-algorithms for object detection \cite{girshick2015region-based} and semantic segmentation \cite{shelhamer2016fully} that take the base architecture as an argument.
In this way GoogLeNet \cite{szegedy2015going} and VGG \cite{szegedy2015going} improved accuracy on a variety of visual problems.
Their patterned components prefigured a more systematic approach to architecture.

Systematic design has delivered deeper and wider networks such as residual networks (ResNets) \cite{he2016deep} and highway networks \cite{srivastava2015highway} for depth and ResNeXT \cite{xie2016aggregated} and FractalNet \cite{larsson2016fractalnet} for width.
While these architectures all contribute their own structural ideas, they incorporated bottlenecks and shortened paths inspired by earlier techniques.
Network-in-network \cite{lin2014network} demonstrated channel mixing as a technique to fuse features, control dimensionality, and go deeper.
The companion and auxiliary losses of deeply-supervised networks \cite{lee2015deeply} and GoogLeNet \cite{szegedy2015going} showed that it helps to keep learned layers and losses close.
For the most part these architectures derive from innovations in connectivity: skipping, gating, branching, and aggregating. 

Our aggregation architectures are most closely related to leading approaches for fusing feature hierarchies.
The key axes of fusion are semantic and spatial. 
Semantic fusion, or aggregating across channels and depths, improves inference of what. 
Spatial fusion, or aggregating across resolutions and scales, improves inference of where.
Deep layer aggregation can be seen as the union of both forms of fusion.

Densely connected networks (DenseNets) \cite{huang2016densely} are the dominant family of architectures for semantic fusion, designed to better propagate features and losses through skip connections that concatenate all the layers in stages.
Our hierarchical deep aggregation shares the same insight on the importance of short paths and re-use, and extends skip connections with trees that cross stages and deeper fusion than concatenation.
Densely connected and deeply aggregated networks achieve more accuracy as well as better parameter and memory efficiency.


Feature pyramid networks (FPNs) \cite{lin2016feature} are the dominant family of architectures for spatial fusion, designed to equalize resolution and standardize semantics across the levels of a pyramidal feature hierarchy through top-down and lateral connections.
Our iterative deep aggregation likewise raises resolution, but further deepens the representation by non-linear and progressive fusion.
FPN connections are linear and earlier levels are not aggregated more to counter their relative semantic weakness.
Pyramidal and deeply aggregated networks are better able to resolve what and where for structured output tasks.


\section{Deep Layer Aggregation}

\begin{figure*}[tb]
    \includegraphics[width=0.9\linewidth]{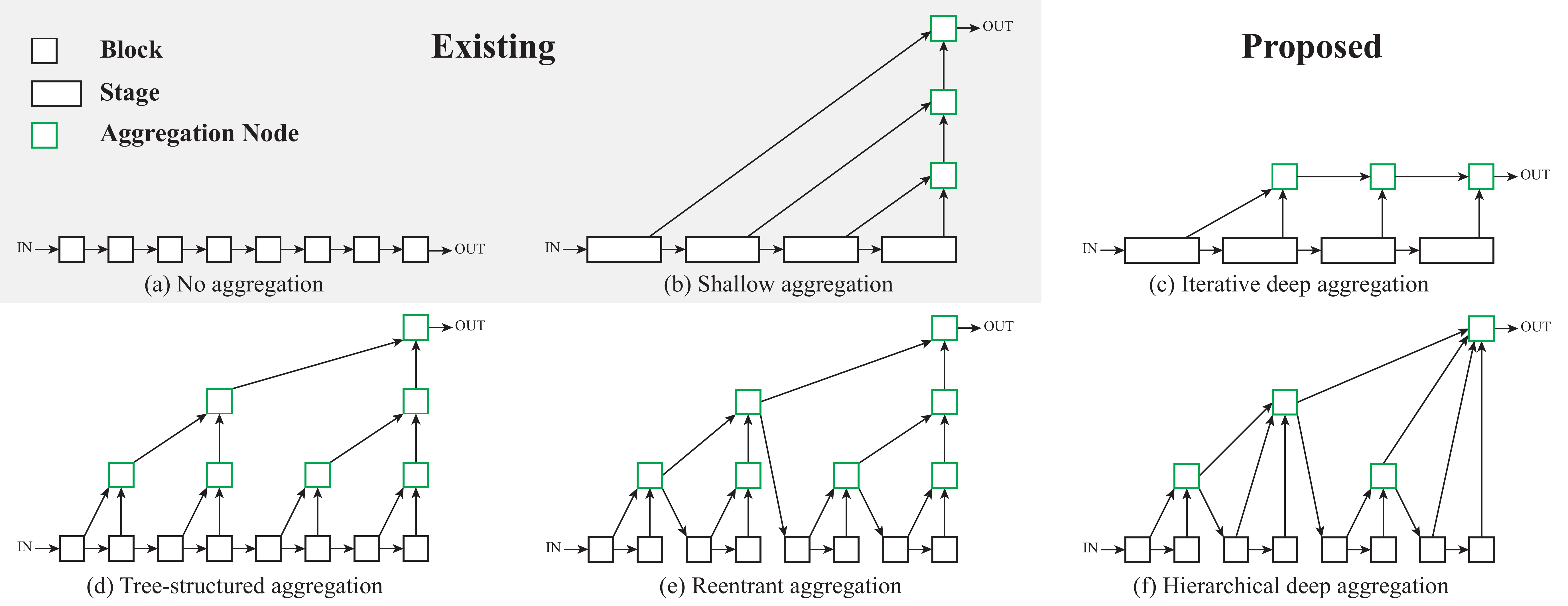}
	\caption{Different approaches to aggregation.
		(a) composes blocks without aggregation as is the default for classification and regression networks.
		(b) combines parts of the network with skip connections, as is commonly used for tasks like segmentation and detection, but does so only shallowly by merging earlier parts in a single step each.
		We propose two deep aggregation architectures: 
		(c) aggregates iteratively by reordering the skip connections of (b) such that the shallowest parts are aggregated the most for further processing and
		(d) aggregates hierarchically through a tree structure of blocks to better span the feature hierarchy of the network across different depths.
		(e) and (f) are refinements of (d) that deepen aggregation by routing intermediate aggregations back into the network and improve efficiency by merging successive aggregations at the same depth.
		Our experiments show the advantages of (c) and (f) for recognition and resolution.}
	\label{fig:struct}
\end{figure*}

We define aggregation as the combination of different layers throughout a network.  
In this work we focus on a family of architectures for the effective aggregation of depths, resolutions, and scales.
We call a group of aggregations {\it deep} if it is compositional, nonlinear, and the earliest aggregated layer passes through multiple aggregations.

As networks can contain many layers and connections, modular design helps counter complexity by grouping and repetition.
Layers are grouped into blocks, which are then grouped into stages by their feature resolution.
We are concerned with aggregating the blocks and stages.


\subsection{Iterative Deep Aggregation}
\label{sec:ida}

Iterative deep aggregation follows the iterated stacking of the backbone architecture.
We divide the stacked blocks of the network into stages according to feature resolution.
Deeper stages are more semantic but spatially coarser.
Skip connections from shallower to deeper stages merge scales and resolutions.
However, the skips in existing work, e.g. FCN~\cite{shelhamer2016fully}, U-Net~\cite{ronneberger2015u}, and FPN~\cite{lin2016feature}, are linear and aggregate the shallowest layers the least, as shown in Figure~\ref{fig:struct}(b).

We propose to instead progressively aggregate and deepen the representation with IDA. 
Aggregation begins at the shallowest, smallest scale and then iteratively merges deeper, larger scales.
In this way shallow features are refined as they are propagated through different stages of aggregation.
Figure~\ref{fig:struct}(c) shows the structure of IDA.

The iterative deep aggregation function $I$ for a series of layers $\vv{x}_1,...,\vv{x}_n$ with increasingly deeper and semantic information is formulated as
\begin{equation}
	\label{eq:lda}
	I(\vv{x}_1, ..., \vv{x}_n) = 
	\begin{cases} 
		\vv{x}_1 & \text{if } n = 1 \\
		I(N(\vv{x}_1, \vv{x}_2), ..., \vv{x}_n) & \text{otherwise},  
	\end{cases}
\end{equation}
where $N$ is the aggregation node.

\subsection{Hierarchical Deep Aggregation}
\label{sec:hda}

Hierarchical deep aggregation merges blocks and stages in a tree to preserve and combine feature channels.
With HDA shallower and deeper layers are combined to learn richer combinations that span more of the feature hierarchy.
While IDA effectively combines stages, it is insufficient for fusing the many blocks of a network, as it is still only sequential.
The deep, branching structure of hierarchical aggregation is shown in Figure~\ref{fig:struct}(d).

Having established the general structure of HDA we can improve its depth and efficiency.
Rather than only routing intermediate aggregations further up the tree, we instead feed the output of an aggregation node back into the backbone as the input to the next sub-tree, as shown in Figure~\ref{fig:struct}(e).
This propagates the aggregation of all previous blocks instead of the preceding block alone to better preserve features.
For efficiency, we merge aggregation nodes of the same depth (combining the parent and left child), as shown in Figure~\ref{fig:struct}(f).

The hierarchical deep aggregation function $T_n$, with depth $n$, is formulated as 
\begin{equation}
	\label{eq:hda}
	\begin{split}
	T_n(\vv{x}) = N(&R^n_{n-1}(\vv{x}), R^n_{n-2}(\vv{x}), ..., \\
	&R^n_1(\vv{x}), L^n_1(\vv{x}), L^n_2(\vv{x})),
	\end{split}
\end{equation}
where $N$ is the aggregation node. $R$ and $L$ are defined as
\begin{equation*}
	\begin{split}
		L^{n}_2(\vv{x}) &= B(L^{n}_1(\vv{x})), \quad L^{n}_1(\vv{x}) = B(R^{n}_1(\vv{x})), \\
		R^n_m(\vv{x}) &= 
		\begin{cases} 
			T_m(\vv{x}) & \text{if } m = n - 1 \\
			T_m(R^n_{m+1}(\vv{x})) & \text{otherwise},
		\end{cases}
	\end{split}
\end{equation*}
where $B$ represents a convolutional block.

\begin{figure*}[tb]
\centering
\includegraphics[width=\linewidth]{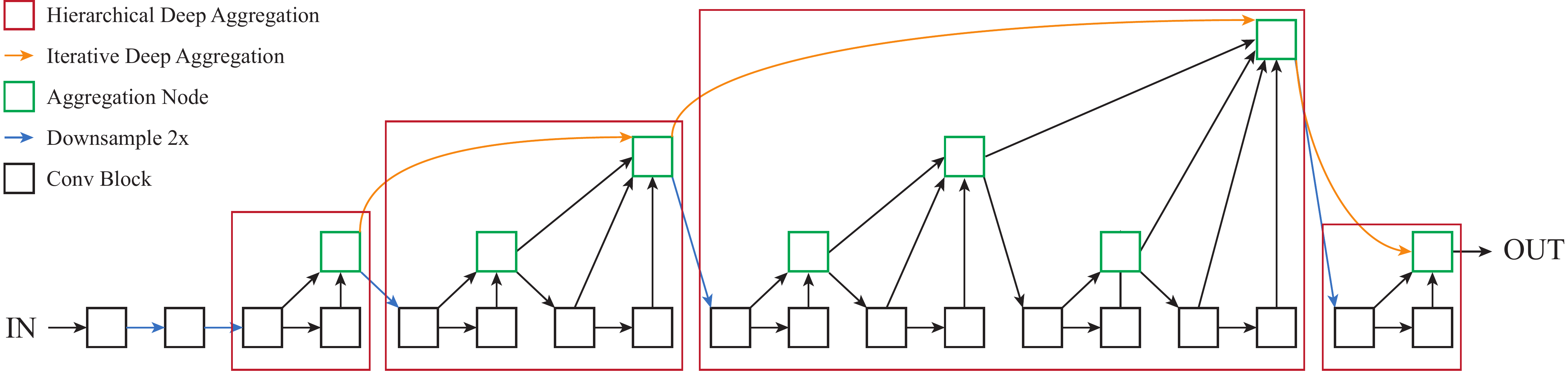}
\caption{
Deep layer aggregation learns to better extract the full spectrum of semantic and spatial information from a network.
Iterative connections join neighboring stages to progressively deepen and spatially refine the representation.
Hierarchical connections cross stages with trees that span the spectrum of layers to better propagate features and gradients.
}
\label{fig:dla60}
\end{figure*}

\subsection{Architectural Elements}
\paragraph{Aggregation Nodes}
The main function of an aggregation node is to combine and compress their inputs.
The nodes learn to select and project important information to maintain the same dimension at their output as a single input.
In our architectures IDA nodes are always binary, while HDA nodes have a variable number of arguments depending on the depth of the tree.

Although an aggregation node can be based on any block or layer, for simplicity and efficiency we choose a single convolution followed by batch normalization and a nonlinearity.
This avoids overhead for aggregation structures.
In image classification networks, all the nodes use 1$\times$1 convolution.
In semantic segmentation, we add a further level of iterative deep aggregation to interpolate features, and in this case use 3$\times$3 convolution.

As residual connections are important for assembling very deep networks, we can also include residual connections in our aggregation nodes.
However, it is not immediately clear that they are necessary for aggregation.
With HDA, the shortest path from any block to the root is at most the depth of the hierarchy, so diminishing or exploding gradients may not appear along the aggregation paths.
In our experiments, we find that residual connection in node can help HDA when the deepest hierarchy has 4 levels or more, while it may hurt for networks with smaller hierarchy.
Our base aggregation, i.e. $N$ in Equation~\ref{eq:lda} and~\ref{eq:hda}, is defined by:

\begin{equation}
N(\vv{x}_1,..., \vv{x}_n) = \sigma(\small{\mathrm{BatchNorm}}(\sum_i W_i \vv{x}_i + \vv{b})),
\end{equation}
where $\sigma$ is the non-linear activation, and $\vv{w}_i$ and $\vv{b}$ are the weights in the convolution. If residual connections are added, the equation becomes
\begin{equation}
	N(\vv{x}_1,..., \vv{x}_n) = \sigma(\small{\mathrm{BatchNorm}}(\sum_i W_i \vv{x}_i + \vv{b}) + \vv{x}_n).
\end{equation}
Note that the order of arguments for $N$ does matter and should follow Equation~\ref{eq:hda}.


\paragraph{Blocks and Stages}
Deep layer aggregation is a general architecture family in the sense that it is compatible with different backbones.
Our architectures make no requirements of the internal structure of the blocks and stages.

The networks we instantiate in our experiments make use of three types of residual blocks \cite{he2016identity, xie2016aggregated}.
Basic blocks combine stacked convolutions with an identity skip connection. 
Bottleneck blocks regularize the convolutional stack by reducing dimensionality through a 1$\times$1 convolution.
Split blocks diversify features by grouping channels into a number of separate paths (called the cardinality of the split).
In this work, we reduce the ratio between the number of output and intermediate channels by half for both bottleneck and split blocks, and the cardinality of our split blocks is 32.
Refer to the cited papers for the exact details of these blocks.

	
\section{Applications}

We now design networks with deep layer aggregation for visual recognition tasks.
To study the contribution of the aggregated representation, we focus on linear prediction without further machinery. 
Our results do without ensembles for recognition and context modeling or dilation for resolution.
Aggregation of semantic and spatial information matters for classification and dense prediction alike.

\subsection{Classification Networks}

Our classification networks augment ResNet and ResNeXT with IDA and HDA.
These are staged networks, which group blocks by spatial resolution, with residual connections within each block.
The end of every stage halves resolution, giving six stages in total, with the first stage maintaining the input resolution while the last stage is 32$\times$ downsampled.
The final feature maps are collapsed by global average pooling then linearly scored.
The classification is predicted as the softmax over the scores.

We connect across stages with IDA and within and across stages by HDA. 
These types of aggregation are easily combined by sharing aggregation nodes.
In this case, we only need to change the root node at each hierarchy by combining Equation~\ref{eq:lda} and~\ref{eq:hda}.
Our stages are downsampled by max pooling with size 2 and stride 2.

The earliest stages have their own structure.
As in DRN~\cite{yu2017dilated}, we replace max pooling in stages 1--2 with strided convolution.
The stage 1 is composed of a 7$\times$7 convolution followed by a basic block.
The stage 2 is only a basic block. 
For all other stages, we make use of combined IDA and HDA on the backbone blocks and stages.

For a direct comparison of layers and parameters in different networks, we build networks with a comparable number of layers as ResNet-34, ResNet-50 and ResNet-101. 
(The exact depth varies as to keep our novel hierarchical structure intact.)
To further illustrate the efficiency of DLA for condensing the representation, we make compact networks with fewer parameters.
Table~\ref{tab:net} lists our networks and Figure~\ref{fig:dla60} shows a DLA architecture with HDA and IDA.

\subsection{Dense Prediction Networks}

\begin{figure}[tb]
\centering
\includegraphics[width=\linewidth]{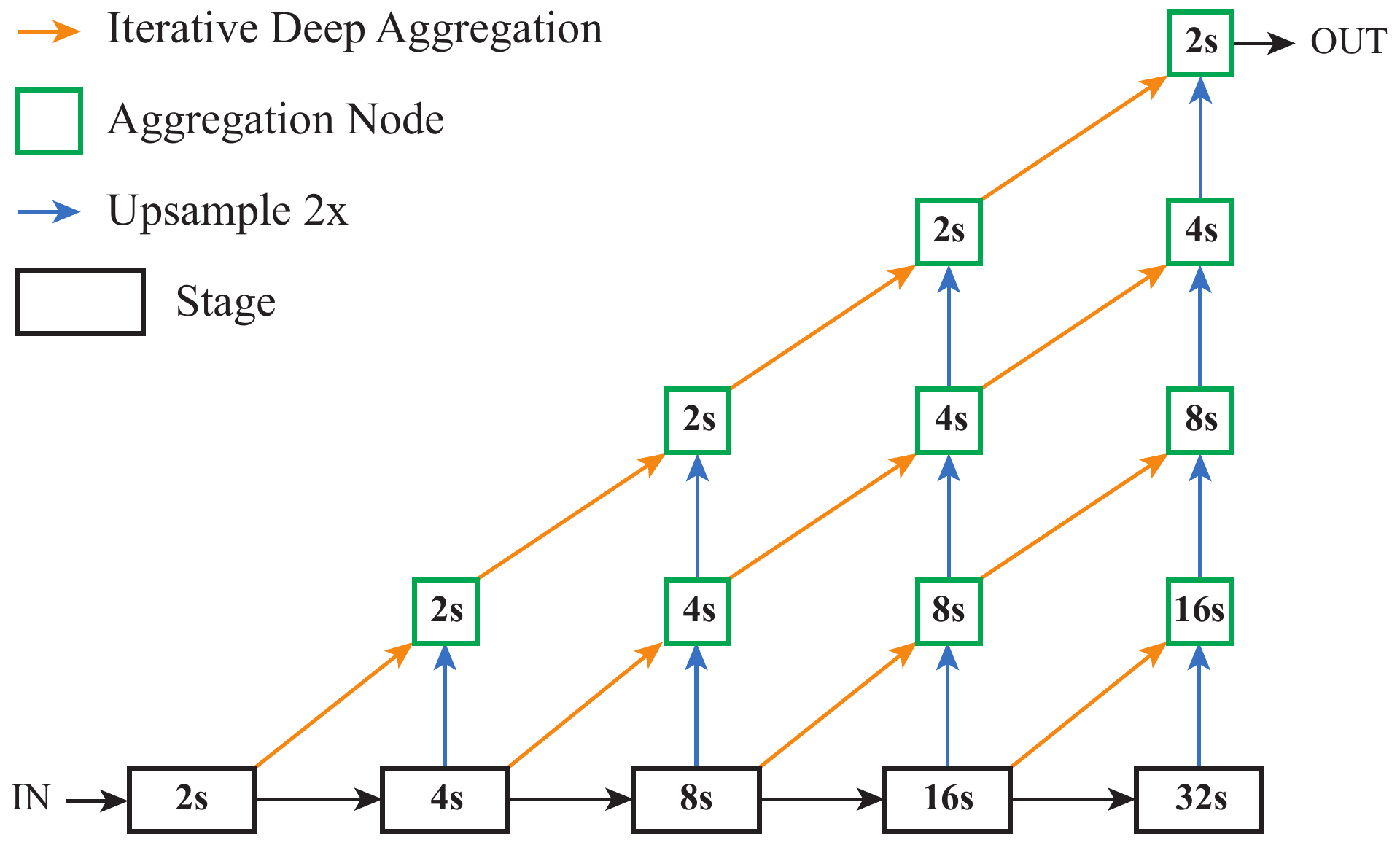}
\caption{
Interpolation by iterative deep aggregation.
Stages are fused from shallow to deep to make a progressively deeper and higher resolution decoder.
}
\label{fig:dense}
\end{figure}

Semantic segmentation, contour detection, and other image-to-image tasks can exploit the aggregation to fuse local and global information.
The conversion from classification DLA to fully convolutional DLA is simple and no different than for other architectures.
We make use of interpolation and a further augmentation with IDA to reach the necessary output resolution for a task.

IDA for interpolation increases both depth and resolution by projection and upsampling as in Figure~\ref{fig:dense}.
All the projection and upsampling parameters are learned jointly during the optimization of the network.
The upsampling steps are initialized to bilinear interpolation and can then be learned as in~\cite{shelhamer2016fully}.
We first project the outputs of stages 3--6 to 32 channels and then interpolate the stages to the same resolution as stage 2.
Finally, we iteratively aggregate these stages to learn a deep fusion of low and high level features.
While having the same purpose as FCN skip connections~\cite{shelhamer2016fully}, hypercolumn features~\cite{hariharan2015hypercolumns}, and FPN top-down connections~\cite{lin2016feature}, our aggregation differs in approach by going from shallow-to-deep to further refine features.
Note that we use IDA twice in this case: once to connect stages in the backbone network and again to recover resolution.

\begin{table*}[t!]
	\adjustbox{max width=\linewidth}{
		\begin{tabular}{l  l  llllll }
			Name & Block & Stage 1 & Stage 2 & Stage 3 & Stage 4 & Stage 5 & Stage 6 \\ \hline
			DLA-34 & Basic & 16 & 32 & 1-64 & 2-128 & 2-256 & 1-512 \\
			DLA-46-C & Bottleneck & 16 & 32 & 1-64 & 2-64 & 2-128 & 1-256 \\
			DLA-60 & Bottleneck & 16 & 32 & 1-128 & 2-256 & 3-512 & 1-1024 \\
			DLA-102 & Bottleneck & 16 & 32 & 1-128 & 3-256 & 4-512 & 1-1024 \\
			DLA-169 & Bottleneck & 16 & 32 & 2-128 & 3-256 & 5-512 & 1-1024 \\ \hline
			DLA-X-46-C & Split & 16 & 32 & 1-64 & 2-64 & 2-128 & 1-256 \\
			DLA-X-60-C & Split & 16 & 32 & 1-64 & 2-64 & 3-128 & 1-256 \\
			DLA-X-60 & Split & 16 & 32 & 1-128 & 2-256 & 3-512 & 1-1024 \\
			DLA-X-102 & Split & 16 & 32 & 1-128 & 3-256 & 4-512 & 1-1024 \\ 
			\hline
		\end{tabular}}
		\caption{
		Deep layer aggregation networks for classification.
		Stages 1 and 2 show the number of channels {\it n} while further stages show {\it d-n} where d is the aggregation depth.
		Models marked with ``-C'' are compact and only have $\sim$1 million parameters.}
		\label{tab:net}
\end{table*}
	
\section{Results}

We evaluate our deep layer aggregation networks on a variety of tasks: image classification on ILSVRC, several kinds of fine-grained recognition, and dense prediction for semantic segmentation and contour detection.
After establishing our classification architecture, we transfer these networks to the other tasks with little to no modification.
DLA improves on or rivals the results of special-purpose networks.

\subsection{ImageNet Classification}
We first train our networks on the ImageNet 2012 training set~\cite{russakovsky2015imagenet}.
Similar to ResNet~\cite{he2016deep}, training is performed by SGD for 120 epochs with momentum 0.9, weight decay $10^{-4}$ and batch size 256.
We start the training with learning rate 0.1, which is reduced by 10 every 30 epochs.
We use scale and aspect ratio augmentation~\cite{xie2016aggregated} with color perturbation.
For fair comparison, we also train the ResNet models with the same training procedure.
This leads to slight improvements over the original results.

We evaluate the performance of trained models on the ImageNet 2012 validation set.
The images are resized so that the shorter side has 256 pixels.
Then central 224$\times$224 crops are extracted from the images and fed into networks to measure prediction accuracy.

\begin{figure*}[htp]
        \centering
        \includegraphics[width=\linewidth]{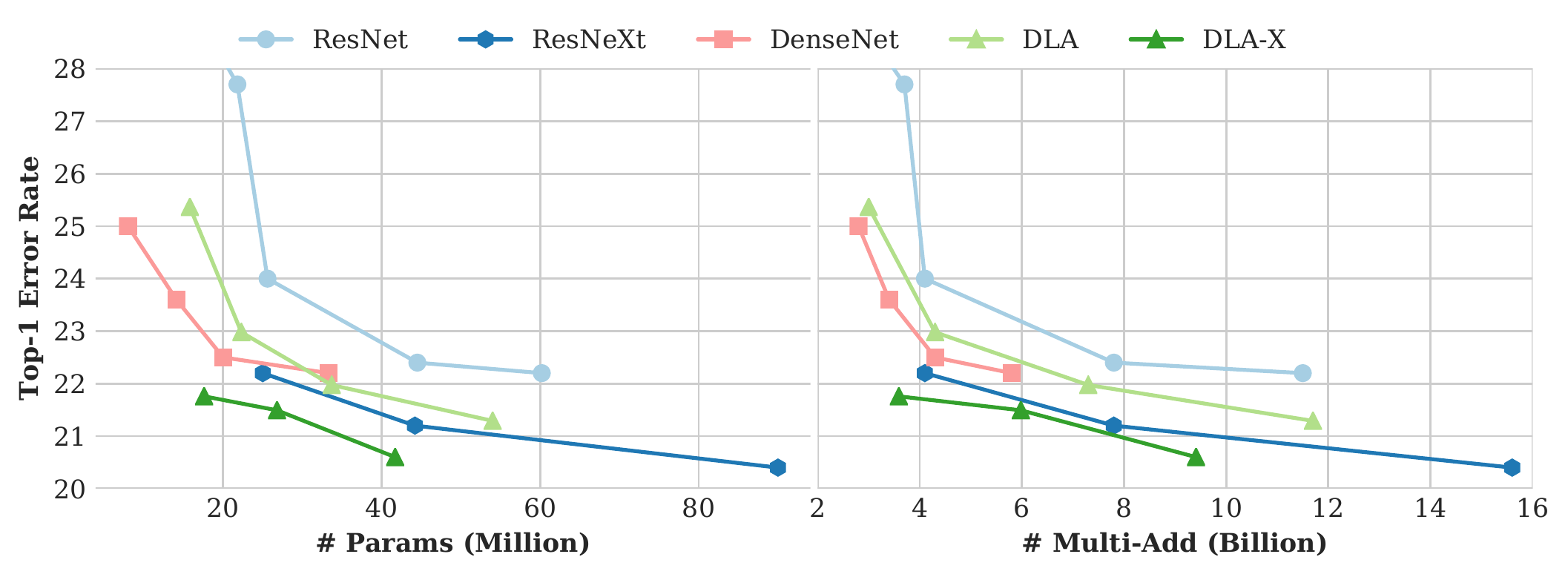}
    
        \caption{
        Evaluation of DLA on ILSVRC.
        DLA/DLA-X have ResNet/ResNeXt backbones respectively.
        DLA achieves the highest accuracies with fewer parameters and fewer computation.
        }
        \label{fig:imagenet_params}
    \end{figure*}

{\bf DLA vs. ResNet} compares DLA networks to ResNets with similar numbers of layers and the same convolutional blocks as shown in Figure~\ref{fig:imagenet_params}.
We find that DLA networks can achieve better performance with fewer parameters.
DLA-34 and ResNet-34 both use basic blocks, but DLA-34 has about 30\% fewer parameters and $\sim1$ point of improvement in top-1 error rate.
We usually expect diminishing returns of performance of deeper networks.
However, our results show that compared to ResNet-50 and ResNet-101, DLA networks can still outperform the baselines significantly with fewer parameters.

{\bf DLA vs. ResNeXt} shows that DLA is flexible enough to use different convolutional blocks and still have advantage in accuracy and parameter efficiency as shown in Figure~\ref{fig:imagenet_params}.
Our models based on the split blocks have much fewer parameters but they still have similar performance with ResNeXt models.
For example, DLA-X-102 has nearly the half number of parameters compared to ResNeXt-101, but the error rate difference is only 0.2\%.

{\bf DLA vs. DenseNet} compares DLA with the dominant architecture for semantic fusion and feature re-use. 
DenseNets are composed of dense blocks that aggregate all of their layers by concatenation and transition blocks that reduce dimensionality for tractability.
While these networks can aggressively reduce depth and parameter count by feature re-use, concatenation is a memory-intensive fusion operation.  
DLA achieves higher accuracy with lower memory usage because the aggregation node fan-in size is log of the total number of convolutional blocks in HDA.

{\bf Compact models} have received a lot of attention due to the limited capabilities of consumer hardware for running convolutional networks.
We design parameter constrained DLA networks to study how efficiently DLA can aggregate and re-use features. We remove color perturbation and set the minimum cropping area to be 0.25 because small models do not have enough capacity for the aggressive data augmentation.
We compare to SqueezeNet~\cite{iandola2016squeezenet}, which shares a block design similar to our own.
Table~\ref{tab:small_model} shows that DLA is more accurate with the same number of parameters.
Furthermore DLA is more computationally efficient by operation count.

\begin{table}[htp]
  \adjustbox{max width=\linewidth}{
\begin{tabular}{l  c  c  c c}
 & Top-1 & Top-5 & Params & FMAs \\ \hline
SqueezNet-A & 42.5 & 19.7 & 1.2M & 1.70B \\
SqueezNet-B & 39.6 & 17.5 & 1.2M & 0.72B\\
DLA-46-C & 35.1 & 13.3 & 1.3M & 0.58B\\
DLA-X-46-C & 34.0 & 13.0 & {\bf 1.1M} & {\bf 0.53B} \\
DLA-X-60-C & {\bf 32.0} & {\bf 11.6} & 1.3M & 0.59B\\
\hline 
\end{tabular}}
\caption{
Comparison with compact models.
DLA is more accurate at the same number of parameters while inference takes fewer operations (counted by fused multiply-adds).
}
\label{tab:small_model}
\end{table}

\subsection{Fine-grained Recognition}

\begin{figure*}[htb]
	\centering
	\includegraphics[width=\linewidth]{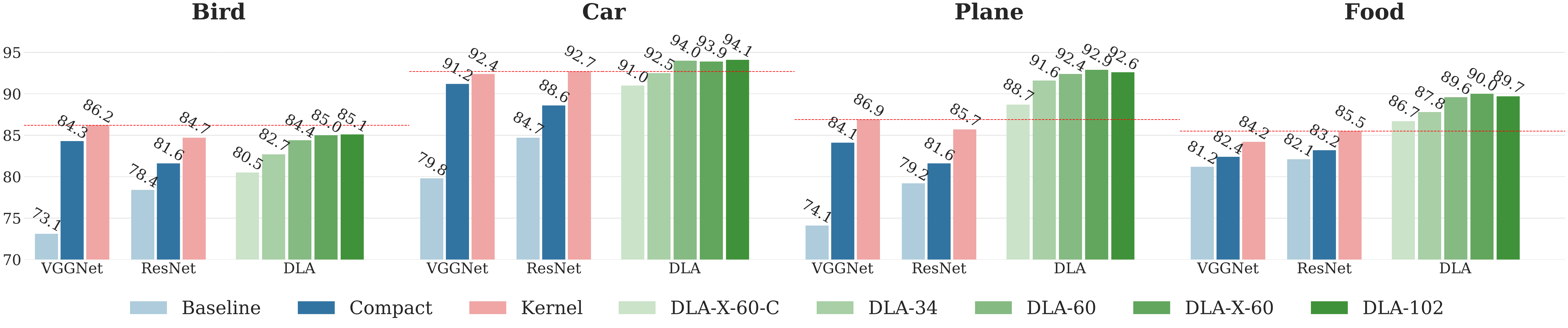}
	\caption{Comparison with state-of-the-art methods on fine-grained datasets. Image classification accuracy on Bird~\cite{wah2011caltech}, Car~\cite{krause20133d}, Plane~\cite{maji2013fine}, and Food~\cite{bossard2014food}. Higher is better. $P$ is the number of parameters in each model. For fair comparison, we calculate the number of parameters for $1000$-way classification. V- and R- indicate the base model as VGGNet-16 and ResNet-50, respectively. The numbers of Baseline, Compact \cite{gao2016compact} and Kernel \cite{cui2017kernel} are directly cited from \cite{cui2017kernel}.}
\label{fig:fine_grained}
\end{figure*}

We use the same training procedure for all of fine-grained experiments.
The training is performed by SGD with a mini-batch size of 64, while the learning rate starts from 0.01 and is then divided by 10 every 50 epochs, for 110 epochs in total.
The other hyperparameters are fixed to their settings for ImageNet classification.
In order to mitigate over-fitting, we carry out the following data augmentation: Inception-style scale and aspect ratio variation \cite{szegedy2015going}, AlexNet-style PCA color noise\cite{krizhevsky2012imagenet}, and the photometric distortions of \cite{howard2013some}.

We evaluate our models on various fine-grained recognition datasets: Bird (CUB)~\cite{wah2011caltech}, Car~\cite{krause20133d}, Plane~\cite{maji2013fine}, and Food~\cite{bossard2014food}.
The statistics of these datasets can be found in Table \ref{tab:fine_grained_datasets}, while results are shown in Figure \ref{fig:fine_grained}.
For fair comparison, we follow the experimental setup of \cite{cui2017kernel}:
we randomly crop 448$\times$448 in resized 512$\times$512 for all the datasets, while keeping 224$\times$224 input size for original VGGNet. 

Our results improve or rival the state-of-the-art without further annotations or specific modules for fine-grained recognition.
In particular, we establish new state-of-the-arts results on Car, Plane, and Food datasets.
Furthermore, our models are competitive while having only several million parameters.
However, our results are not better than state-of-the-art on Birds, although note that this dataset has fewer instances per class so further regularization might help.

\begin{table}[ht]
  \centering
  \adjustbox{max width=\columnwidth}{
		\begin{tabular}{l c c c}
			 & \#Class & \#Train (per class) & \#Test (per class) \\
			\hline 
            Bird& 200 & 5994 (30) & 5794 (29) \\
            Car & 196 & 8144 (42) & 8041 (41) \\
            Plane & 102 & 6667 (67) & 3333 (33)\\
            Food & 101 & 75750 (750) & 25250 (250) \\
			\hline 
            ILSVRC & 1000 & 1,281,167 (1281) & 100,000 (100) \\
            \hline 
		\end{tabular}}
	\caption{
	Statistics for fine-grained recognition datasets. 
	Compared to generic, large-scale classification, these tasks contain more specific classes with fewer training instances. 
	}
	\label{tab:fine_grained_datasets}
\end{table}

\begin{table*}[htb]
\begin{floatrow}[2]
\tablebox{\caption{
Evaluation on Cityscapes to compare strides on validation and to compare against existing methods on test.
DLA is the best-in-class among methods in the same setting.
}}{
 \adjustbox{max width=\linewidth}{
    \begin{tabular}{ccc}
        Method & Split & mIoU \\ \hline
        DLA-34 8s & \multirow{3}{*}{Val} & 73.5 \\
        DLA-34 2s & & 75.1 \\
        DLA-102 2s & & 74.4 \\
        \hline
        FCN-8s \cite{shelhamer2016fully} & \multirow{4}{*}{Test} & 65.3 \\
        RefineNet-101 \cite{lin2016refinenet} & & 73.6 \\
        DLA-102 & & 75.3 \\
        DLA-169 & & {\bf 75.9} \\ \hline
    \end{tabular}}
\label{tab:cityscapes}
}
\tablebox{\caption{
Evaluation on CamVid.
Higher depth and resolution help.
DLA is state-of-the-art.
}}{
  \adjustbox{max width=\linewidth}{
\begin{tabular}{cc}
        Method & mIoU \\ \hline
        SegNet \cite{badrinarayanan2015segnet} & 46.4 \\
        DeepLab-LFOV \cite{chen2014semantic} & 61.6 \\
        Dilation8 \cite{yu2016multi} & 65.3 \\
        FSO \cite{kundu2016feature} & 66.1 \\ \hline
        DLA-34 8s & 66.7 \\
        DLA-34 2s & 68.6 \\
        DLA-102 2s & {\bf 71.0} \\ \hline
    \end{tabular}}
\label{tab:camvid}
}
\end{floatrow}
\end{table*}


\subsection{Semantic Segmentation}

We report experiments for urban scene understanding on CamVid~\cite{brostow2009semantic} and Cityscapes~\cite{cordts2016cityscapes}.
Cityscapes is a larger-scale, more challenging dataset for comparison with other methods while CamVid is more convenient for examining ablations.
We use the standard mean intersection-over-union (IoU) score~\cite{everingham2010pascal} as the evaluation metric for both datasets.
Our networks are trained only on the training set without the usage of validation or other further data.


CamVid has $367$ training images, $100$ validation images, and $233$ test images with $11$ semantic categories.
We start the training with learning rate 0.01 and divide it by 10 after 800 epochs.
The results are shown in Table~\ref{tab:camvid}.
We find that models with downsampling rate 2 consistenly outperforms those downsampling by 8.
We also try to augment the data by randomly rotating the images between [-10, 10] degrees and randomly scaling the images between 0.5 and 2. 
The final results are significantly better than prior methods.


Cityscapes has $2,975$ training images, $500$ validation images, and $1,525$ test images with $19$ semantic categories.
Following previous works~\cite{zhao2016pyramid}, we adopt the poly learning rate $(1 - \frac{\text{epoch}-1}{\text{total epoch}})^{0.9}$ with momentum 0.9 and train the model for 500 epochs with batch size 16.
The starting learning rate is 0.01 and the crop size is chosen to be 864.
We also augment the data by randomly rotating within 10 degrees and scaling between 0.5 and 2.
The validation results are shown in~\ref{tab:cityscapes}.
Surprisingly, DLA-34 performs very well on this dataset and it is as accurate as DLA-102.
It should be noted that fine spatial details do not contribute much for this choice of metric.
RefineNet~\cite{lin2016refinenet} is the strongest network in the same class of methods without the computational costs of additional data, dilation, and graphical models.
To make a fair comparison, we evaluate in the same multi-scale fashion as that approach with image scales of [0.5, 0.75, 1, 1.25, 1.5] and sum the predictions.
DLA improves by $2+$ points.


\subsection{Boundary Detection}

\begin{table}[htp]
    \centering
    \begin{tabular}{l ccc}
        Method & ODS & OIS & AP \\ \midrule
        SE~\cite{dollar2013structured} & 0.746 & 0.767 & 0.803 \\
        DeepEdge~\cite{bertasius2015deepedge} & 0.753 & 0.772 & 0.807 \\
        DeepContour~\cite{shen2015deepcontour} & 0.756 &  0.773 & 0.797 \\
        HED~\cite{xie2015holistically} & 0.788 & 0.808 & 0.840 \\
        CEDN~\cite{yang2016object}$^{\dagger}$ & 0.788 & 0.804 & 0.821 \\
        UberNet~\cite{kokkinos2016ubernet} (1-Task)$^{\dagger}$ & 0.791 & 0.809 & {\bf 0.849} \\ \midrule
        DLA-34 8s & 0.760 & 0.772 & 0.739 \\
        DLA-34 4s & 0.767 & 0.778 & 0.751 \\
        DLA-34 2s & 0.794 & 0.808 & 0.787 \\
        DLA-102 2s & {\bf 0.803} & {\bf 0.813} & 0.781 \\
    \midrule
    \end{tabular}
    \caption{
    Evaluation on BSDS ($^{\dagger}$ indicates outside data).
    ODS and OIS are state-of-the-art, but AP suffers due to recall.
    }
    \label{tab:edge-bsds}
\end{table}

\begin{table}[htp]
    \centering
    \begin{tabular}{lcccc}
        Method & Train & ODS & OIS & AP \\ \midrule
        SE~\cite{dollar2013structured} & \multirow{4}{*}{BSDS} & 0.541 & 0.570  & 0.486 \\
        HED~\cite{yang2016object} & &  0.553 & 0.585 & 0.518 \\
        DLA-34 2s & & 0.642 & 0.668 & 0.624 \\
        DLA-102 2s & & 0.648 & 0.674 & 0.623 \\
        \midrule
        DSBD~\cite{premachandran2017pascal} & \multirow{4}{*}{PASCAL} & 0.643 & 0.663 & 0.650 \\
        M-DSBD~\cite{premachandran2017pascal} & & 0.652 & 0.678 & 0.674 \\
        DLA-34 2s & & 0.743 & 0.757 & {\bf 0.763} \\
        DLA-102 2s & & {\bf 0.754} & {\bf 0.766} & 0.752 \\
        \midrule
    \end{tabular}
    \caption{
    Evaluation on PASCAL Boundaries.
    DLA is state-of-the-art.
    }
    \label{tab:edge-pascal}
\end{table}

\begin{figure}
    \centering
    \includegraphics[width=0.9\linewidth]{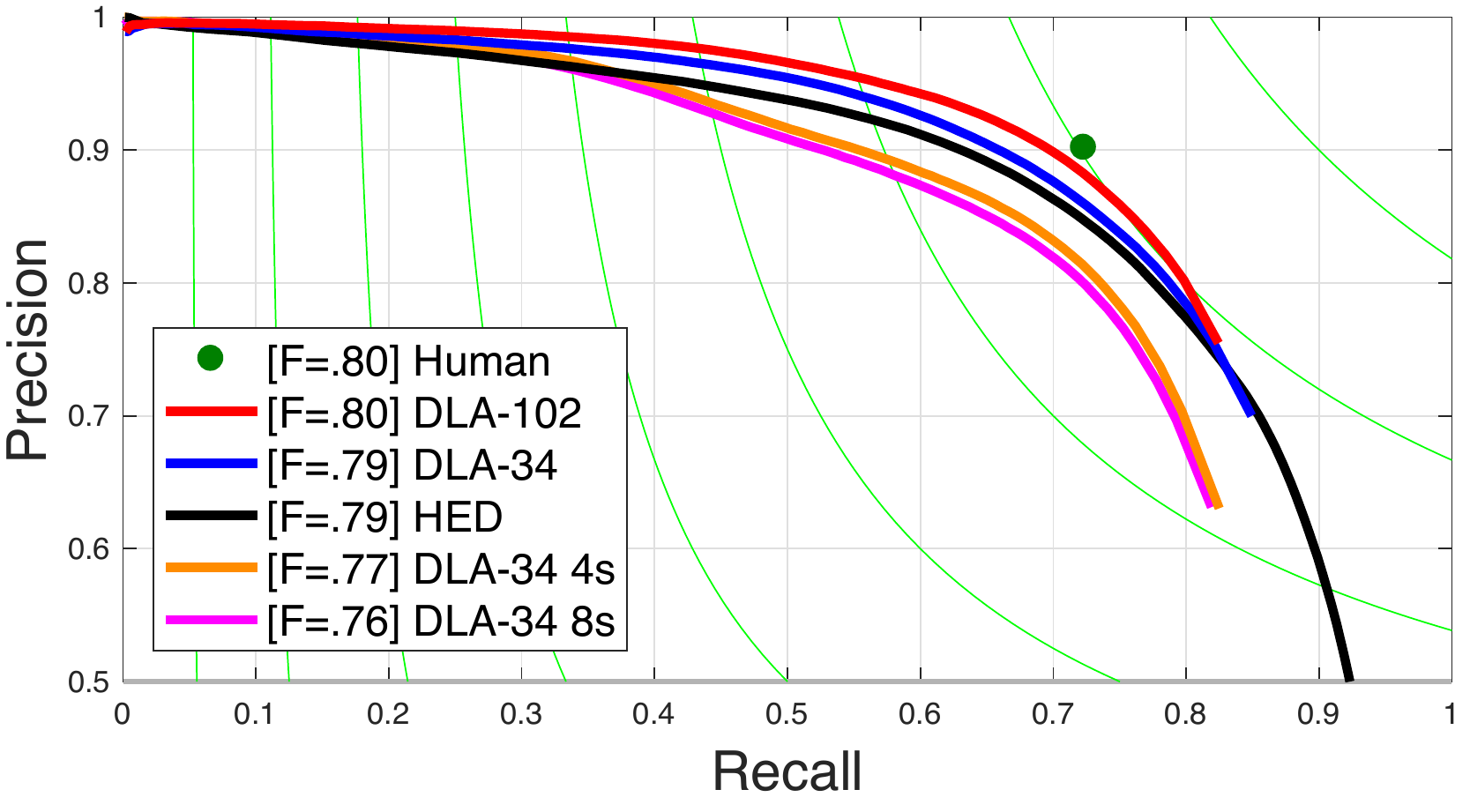}
    \caption{
    Precision-recall evaluation on BSDS.
    DLA is the closest to human performance.
    }
    \label{fig:bsds}
\end{figure}

Boundary detection is an exacting task of localization.
Although as a classification problem it is only a binary task of whether or not a boundary exists, the metrics require precise spatial accuracy.
We evaluate on classic BSDS~\cite{arbelaez2011contour} with multiple human boundary annotations and PASCAL boundaries \cite{premachandran2017pascal} with boundaries defined by instances masks of select semantic classes.
The metrics are accuracies at different thresholds, the optimal dataset scale (ODS) and more lenient optimal image scale (OIS), as well as average precision (AP).
Results are shown in for BSDS in Table~\ref{tab:edge-bsds} and the precision-recall plot of Figure~\ref{fig:bsds} and for PASCAL boundaries in Table~\ref{tab:edge-pascal}.

To address this task we follow the training procedure of HED~\cite{xie2015holistically}.
In line with other deep learning methods, we take the consensus of human annotations on BSDS and only supervise our network with boundaries that three or more annotators agree on. Following~\cite{yang2016object}, we give the boundary labels 10 times weight of the others.
For inference we simply run the net forward, and do not make use of ensembles or multi-scale testing.
Assessing the role of resolution by comparing strides of 8, 4, and 2 we find that high output resolution is critical for accurate boundary detection. We also find that deeper networks does not continue improving the prediction performance on BSDS.

On both BSDS and PASCAL boundaries we achieve state-of-the-art ODS and OIS scores.
In contrast the AP on BSDS is surprisingly low, so to understand why we plot the precision-recall curve in Figure~\ref{fig:bsds}. 
Our approach has lower recall, but this is explained by the consensus ground truth not covering all of the individual, noisy boundaries.
At the same time it is the closest to human performance.
On the other hand we achieve state-of-the-art AP on PASCAL boundaries since it has a single, consistent notion of boundaries.
When training on BSDS and transferring to PASCAL boundaries the improvement is minor, but training on PASCAL boundaries itself with $\sim10\times$ the data delivers more than $10\%$ relative improvement over competing methods.

\section{Conclusion}

Aggregation is a decisive aspect of architecture, and as the number of modules multiply their connectivity is made all the more important.
By relating architectures for aggregating channels, scales, and resolutions we identified the need for deeper aggregation, and addressed it by iterative deep aggregation and hierarchical deep aggregation.
Our models are more accurate and make more efficient use of parameters and computation than baseline networks.
Our aggregation extensions improve on dominant architectures like residual and densely connected networks.
Bridging the gaps of architecture makes better use of layers in aggregate.

{
\bibliographystyle{ieee}
\bibliography{ref}

\begin{thebibliography}{10}\itemsep=-1pt

\bibitem{arbelaez2011contour}
P.~Arbelaez, M.~Maire, C.~Fowlkes, and J.~Malik.
\newblock Contour detection and hierarchical image segmentation.
\newblock {\em TPAMI}, 2011.

\bibitem{badrinarayanan2015segnet}
V.~Badrinarayanan, A.~Kendall, and R.~Cipolla.
\newblock Segnet: A deep convolutional encoder-decoder architecture for image
  segmentation.
\newblock {\em arXiv preprint arXiv:1511.00561}, 2015.

\bibitem{bertasius2015deepedge}
G.~Bertasius, J.~Shi, and L.~Torresani.
\newblock {DeepEdge}: A multi-scale bifurcated deep network for top-down
  contour detection.
\newblock In {\em CVPR}, 2015.

\bibitem{bishop2006pattern}
C.~M. Bishop.
\newblock {\em Pattern recognition and machine learning}, page 229.
\newblock Springer-Verlag New York, 2006.

\bibitem{bossard2014food}
L.~Bossard, M.~Guillaumin, and L.~Van~Gool.
\newblock Food-101--mining discriminative components with random forests.
\newblock In {\em ECCV}, 2014.

\bibitem{brostow2009semantic}
G.~J. Brostow, J.~Fauqueur, and R.~Cipolla.
\newblock Semantic object classes in video: A high-definition ground truth
  database.
\newblock {\em Pattern Recognition Letters}, 2009.

\bibitem{chen2014semantic}
L.-C. Chen, G.~Papandreou, I.~Kokkinos, K.~Murphy, and A.~L. Yuille.
\newblock Semantic image segmentation with deep convolutional nets and fully
  connected crfs.
\newblock In {\em ICLR}, 2015.

\bibitem{cordts2016cityscapes}
M.~Cordts, M.~Omran, S.~Ramos, T.~Rehfeld, M.~Enzweiler, R.~Benenson,
  U.~Franke, S.~Roth, and B.~Schiele.
\newblock The cityscapes dataset for semantic urban scene understanding.
\newblock In {\em CVPR}, 2016.

\bibitem{cui2017kernel}
Y.~Cui, F.~Zhou, J.~Wang, X.~Liu, Y.~Lin, and S.~Belongie.
\newblock Kernel pooling for convolutional neural networks.
\newblock In {\em CVPR}, 2017.

\bibitem{dollar2013structured}
P.~Doll{\'a}r and C.~L. Zitnick.
\newblock Structured forests for fast edge detection.
\newblock In {\em ICCV}, 2013.

\bibitem{donahue2014decaf}
J.~{Donahue}, Y.~{Jia}, O.~{Vinyals}, J.~{Hoffman}, N.~{Zhang}, E.~{Tzeng}, and
  T.~{Darrell}.
\newblock Decaf: A deep convolutional activation feature for generic visual
  recognition.
\newblock In {\em ICML}, 2014.

\bibitem{everingham2010pascal}
M.~Everingham, L.~Van~Gool, C.~K. Williams, J.~Winn, and A.~Zisserman.
\newblock The pascal visual object classes (voc) challenge.
\newblock {\em IJCV}, 2010.

\bibitem{gao2016compact}
Y.~Gao, O.~Beijbom, N.~Zhang, and T.~Darrell.
\newblock Compact bilinear pooling.
\newblock In {\em CVPR}, 2016.

\bibitem{girshick2015region-based}
R.~Girshick, J.~Donahue, T.~Darrell, and J.~Malik.
\newblock Region-based convolutional networks for accurate object detection and
  segmentation.
\newblock {\em TPAMI}, 2015.

\bibitem{hariharan2015hypercolumns}
B.~Hariharan, P.~Arbel{\'a}ez, R.~Girshick, and J.~Malik.
\newblock Hypercolumns for object segmentation and fine-grained localization.
\newblock In {\em CVPR}, 2015.

\bibitem{he2016deep}
K.~He, X.~Zhang, S.~Ren, and J.~Sun.
\newblock Deep residual learning for image recognition.
\newblock In {\em CVPR}, 2016.

\bibitem{he2016identity}
K.~He, X.~Zhang, S.~Ren, and J.~Sun.
\newblock Identity mappings in deep residual networks.
\newblock In {\em ECCV}, 2016.

\bibitem{howard2013some}
A.~G. Howard.
\newblock Some improvements on deep convolutional neural network based image
  classification.
\newblock {\em arXiv preprint arXiv:1312.5402}, 2013.

\bibitem{huang2016densely}
G.~Huang, Z.~Liu, K.~Q. Weinberger, and L.~van~der Maaten.
\newblock Densely connected convolutional networks.
\newblock In {\em CVPR}, 2017.

\bibitem{iandola2016squeezenet}
F.~N. Iandola, S.~Han, M.~W. Moskewicz, K.~Ashraf, W.~J. Dally, and K.~Keutzer.
\newblock Squeezenet: Alexnet-level accuracy with 50x fewer parameters and $<$
  0.5 mb model size.
\newblock {\em arXiv preprint arXiv:1602.07360}, 2016.

\bibitem{kokkinos2016ubernet}
I.~Kokkinos.
\newblock Ubernet: Training auniversal'convolutional neural network for low-,
  mid-, and high-level vision using diverse datasets and limited memory.
\newblock {\em arXiv preprint arXiv:1609.02132}, 2016.

\bibitem{krause20133d}
J.~Krause, M.~Stark, J.~Deng, and L.~Fei-Fei.
\newblock 3d object representations for fine-grained categorization.
\newblock In {\em ICCV Workshops}, 2013.

\bibitem{krizhevsky2012imagenet}
A.~Krizhevsky, I.~Sutskever, and G.~E. Hinton.
\newblock Imagenet classification with deep convolutional neural networks.
\newblock In {\em NIPS}, 2012.

\bibitem{kundu2016feature}
A.~Kundu, V.~Vineet, and V.~Koltun.
\newblock Feature space optimization for semantic video segmentation.
\newblock In {\em CVPR}, 2016.

\bibitem{larsson2016fractalnet}
G.~Larsson, M.~Maire, and G.~Shakhnarovich.
\newblock Fractalnet: Ultra-deep neural networks without residuals.
\newblock In {\em ICLR}, 2017.

\bibitem{lecun1998gradient}
Y.~LeCun, L.~Bottou, Y.~Bengio, and P.~Haffner.
\newblock Gradient-based learning applied to document recognition.
\newblock {\em Proceedings of the IEEE}, 86(11):2278--2324, 1998.

\bibitem{lee2015deeply}
C.-Y. Lee, S.~Xie, P.~Gallagher, Z.~Zhang, and Z.~Tu.
\newblock Deeply-supervised nets.
\newblock In {\em Artificial Intelligence and Statistics}, pages 562--570,
  2015.

\bibitem{lin2016refinenet}
G.~Lin, A.~Milan, C.~Shen, and I.~Reid.
\newblock Refinenet: Multi-path refinement networks with identity mappings for
  high-resolution semantic segmentation.
\newblock In {\em CVPR}, 2017.

\bibitem{lin2014network}
M.~Lin, Q.~Chen, and S.~Yan.
\newblock Network in network.
\newblock In {\em ICLR}, 2014.

\bibitem{lin2016feature}
T.-Y. Lin, P.~Doll{\'a}r, R.~Girshick, K.~He, B.~Hariharan, and S.~Belongie.
\newblock Feature pyramid networks for object detection.
\newblock In {\em CVPR}, 2017.

\bibitem{maji2013fine}
S.~Maji, E.~Rahtu, J.~Kannala, M.~Blaschko, and A.~Vedaldi.
\newblock Fine-grained visual classification of aircraft.
\newblock {\em arXiv preprint arXiv:1306.5151}, 2013.

\bibitem{premachandran2017pascal}
V.~Premachandran, B.~Bonev, X.~Lian, and A.~Yuille.
\newblock Pascal boundaries: A semantic boundary dataset with a deep semantic
  boundary detector.
\newblock In {\em WACV}, 2017.

\bibitem{ronneberger2015u}
O.~Ronneberger, P.~Fischer, and T.~Brox.
\newblock U-net: Convolutional networks for biomedical image segmentation.
\newblock In {\em International Conference on Medical Image Computing and
  Computer-Assisted Intervention}, 2015.

\bibitem{russakovsky2015imagenet}
O.~Russakovsky, J.~Deng, H.~Su, J.~Krause, S.~Satheesh, S.~Ma, Z.~Huang,
  A.~Karpathy, A.~Khosla, M.~Bernstein, et~al.
\newblock Imagenet large scale visual recognition challenge.
\newblock {\em IJCV}, 2015.

\bibitem{shelhamer2016fully}
E.~Shelhamer, J.~Long, and T.~Darrell.
\newblock Fully convolutional networks for semantic segmentation.
\newblock {\em TPAMI}, 2016.

\bibitem{shen2015deepcontour}
W.~Shen, X.~Wang, Y.~Wang, X.~Bai, and Z.~Zhang.
\newblock {DeepContour}: A deep convolutional feature learned by
  positive-sharing loss for contour detection.
\newblock In {\em CVPR}, 2015.

\bibitem{simonyan2015very}
K.~Simonyan and A.~Zisserman.
\newblock Very deep convolutional networks for large-scale image recognition.
\newblock In {\em ICLR}, 2015.

\bibitem{srivastava2015highway}
R.~K. Srivastava, K.~Greff, and J.~Schmidhuber.
\newblock Highway networks.
\newblock In {\em NIPS}, 2015.

\bibitem{szegedy2015going}
C.~Szegedy, W.~Liu, Y.~Jia, P.~Sermanet, S.~Reed, D.~Anguelov, D.~Erhan,
  V.~Vanhoucke, and A.~Rabinovich.
\newblock Going deeper with convolutions.
\newblock In {\em CVPR}, 2015.

\bibitem{wah2011caltech}
C.~Wah, S.~Branson, P.~Welinder, P.~Perona, and S.~Belongie.
\newblock The caltech-ucsd birds-200-2011 dataset.
\newblock 2011.

\bibitem{xie2016aggregated}
S.~Xie, R.~Girshick, P.~Doll{\'a}r, Z.~Tu, and K.~He.
\newblock Aggregated residual transformations for deep neural networks.
\newblock In {\em CVPR}, 2017.

\bibitem{xie2015holistically}
S.~Xie and Z.~Tu.
\newblock Holistically-nested edge detection.
\newblock In {\em ICCV}, 2015.

\bibitem{yang2016object}
J.~Yang, B.~Price, S.~Cohen, H.~Lee, and M.-H. Yang.
\newblock Object contour detection with a fully convolutional encoder-decoder
  network.
\newblock In {\em CVPR}, 2016.

\bibitem{yosinski2014transferable}
J.~Yosinski, J.~Clune, Y.~Bengio, and H.~Lipson.
\newblock How transferable are features in deep neural networks?
\newblock In {\em NIPS}, 2014.

\bibitem{yu2016multi}
F.~Yu and V.~Koltun.
\newblock Multi-scale context aggregation by dilated convolutions.
\newblock In {\em ICLR}, 2016.

\bibitem{yu2017dilated}
F.~Yu, V.~Koltun, and T.~Funkhouser.
\newblock Dilated residual networks.
\newblock In {\em CVPR}, 2017.

\bibitem{zagoruyko2016wide}
S.~Zagoruyko and N.~Komodakis.
\newblock Wide residual networks.
\newblock {\em arXiv preprint arXiv:1605.07146}, 2016.

\bibitem{zeiler2014visualizing}
M.~D. Zeiler and R.~Fergus.
\newblock Visualizing and understanding convolutional networks.
\newblock In {\em ECCV}, 2014.

\bibitem{zhao2016pyramid}
H.~Zhao, J.~Shi, X.~Qi, X.~Wang, and J.~Jia.
\newblock Pyramid scene parsing network.
\newblock {\em arXiv preprint arXiv:1612.01105}, 2016.

\end{thebibliography}
}




\end{document}